\documentclass[runningheads]{llncs}

\usepackage{graphicx}
\usepackage{pythontex}
\usepackage{tabu}
\usepackage{float}
\usepackage{hyperref}

\begin{document}
\title{Imperfect Segmentation Labels:\\ How Much Do They Matter?}
\titlerunning{Imperfect Segmentation Labels: How Much Do They Matter?}

\author{Nicholas Heller \and
Joshua Dean \and
Nikolaos Papanikolopoulos}
\authorrunning{N. Heller et al.}
\institute{Computer Science and Engineering, University of Minnesota -- Twin Cities
\email{\{helle246, deanx252, papan001\}@umn.edu}\\
}
\maketitle 
\begin{abstract}
Labeled datasets for semantic segmentation are imperfect, especially in medical imaging where borders are often subtle or ill-defined. Little work has been done to analyze the effect that label errors have on the performance of segmentation methodologies. Here we present a large-scale study of model performance in the presence of varying types and degrees of error in training data. We trained U-Net, SegNet, and FCN32 several times for liver segmentation with 10 different modes of ground-truth perturbation. Our results show that for each architecture, performance steadily declines with boundary-localized errors, however, U-Net was significantly more robust to jagged boundary errors than the other architectures. We also found that each architecture was very robust to non-boundary-localized errors, suggesting that boundary-localized errors are fundamentally different and more challenging problem than random label errors in a classification setting.

\end{abstract}
\section{Introduction}

Automatic semantic segmentation has wide applications in medicine, including new visualization techniques \cite{DBLP:journals/corr/abs-1803-05431}, surgical simulation \cite{doi:10.1080/17434440.2018.1473033}, and larger studies of morphological features \cite{Chang2005}, all of which would remain prohibitively expensive if segmentations were provided manually.

In the past 4 years, Deep Learning (DL) has risen to the forefront of semantic segmentation techniques with virtually all segmentation challenges currently dominated by DL-based entries \cite{GuoYanming2016Dlfv}. Deep learning is a subfield of machine learning which uses labeled input, or training data to learn functions that map unlabeled input data to its correct response. In the case of semantic segmentation, the model learns from image and mask pairs, where the mask assigns each pixel or voxel to one of a set number of classes. These masks are typically provided manually by a domain expert and often contain some errors.

Typically the most challenging and expensive task in using deep learning for semantic segmentation is curating a ground-truth dataset that is sufficiently large for the trained model to effectively generalize to unseen data. Practitioners are often faced with a tradeoff between the quantity of ground-truth masks and their quality \cite{B2017}.

We categorize ground truth errors to be either \textit{biased} or \textit{unbiased}. Biased errors stem from errors of intention, where the expert creating the labels would repeat the error if asked to label the instance again. These errors are pernicious because they can result in systemic inaccuracies in the dataset that may then be imparted to the learned model. These errors can often be mitigated by giving clear and unambiguous instructions for those performing the labeling. 

Unbiased errors are all other types of errors. For instance, if an annotator's hand shakes when performing labeling, this would be an unbiased error so long as his hand is not more likely to shake on certain features than on others. We define the \textit{gold standard} ground truth to be what an unbiased annotator would produce if he were to annotate every instance an infinite number of times and then take plurality votes to produce the final labels. For semantic segmentation, each pixel would be an instance in this example.

\begin{figure}
    \centering
    \includegraphics[width=11cm]{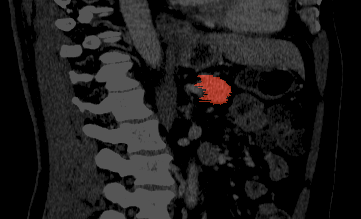}
    \caption{A sagittal cross-section of an annotation from the Pancreas Segmentation Dataset \cite{roth2015deeporgan} that was performed in the axial plane (best viewed in color).}
    \label{fig:dif_plane_view}
\end{figure}

Errors can be difficult to recognize in annotated images, but in medical imaging, 3D imaging modalities such as Computed Tomography (CT) allow us to scrutinize the annotations from the other anatomical planes. In Fig. \ref{fig:dif_plane_view} we can clearly see that the expert is somewhat inconsistent in his treatment of the region boundary in the axial plane, since there are clear discontinuities in the contour when viewed from the saggital plane. This is important in medical image processing because often models are trained on all three anatomical planes to produce a more robust final segmentation \cite{prasoon2013deep}, or volumetric models are used \cite{milletari2016v}. It's conceivable that this jagged boundary might confuse a learned model by suggesting that the predicted segmentations should also have jagged boundaries.

In this work, we study how errors in ground truth masks affect the performance of trained models for the task of semantic segmentation in medical imaging. In particular, we simulate ground truth errors in the widely used Liver Segmentation Dataset\footnote{https://competitions.codalab.org/competitions/17094} by perturbing the training annotations to various degrees in a "natural", "choppy", and "random" way. The validation and testing annotations were left untouched. We repeatedly train three widely used DL-based segmentation models (U-Net \cite{DBLP:journals/corr/RonnebergerFB15}, SegNet \cite{DBLP:journals/corr/BadrinarayananK15}, and FCN32 \cite{DBLP:journals/corr/ShelhamerLD16}) on the perturbed training data and report the corresponding degradation in performance.

\section{Related Work}

In \cite{Angluin1988} Angluin and Laird analyzed mislabeled examples from the standpoint of Probably Approximately Correct learning. They show that the learning problem remains feasible so long as the noise affects less than half of the instances on average, although sample complexity increases with label noise.

Considerable work has been done to characterize the effect of label noise on classical algorithms such as Decision Trees, Support Vector Machines, and k-Nearest Neighbors, and robust variants of these have been proposed. For a detailed survey, see \cite{Frenay2014}. Many data-cleansing algorithms have been proposed to reduce the incidence of mislabeled data in datasets \cite{Muhlenbach2002} \cite{brodley1999identifying} \cite{verbaeten2003ensemble} but challenges arise in distinguishing mislabeled instances from instances that are difficult but informative.

With the rise to prominence of deep learning for computer vision tasks, the ready availability of vast quantities of noisily labeled data on the internet, and the lack of sufficient data-cleansing algorithms, many have turned their attention to studying the pitfalls of training Deep Neural Networks for image recognition, attribute learning, and scene classification using noisy labels. 

In \cite{Xiao2015} the authors find that transfer learning from a noisy dataset to a smaller but clean dataset for the same task does better than fine-tuning on the clean dataset alone. They go on to extend Convolutional Neural Networks with a probabilistic framework to model how mislabelings occur and infer true labels with the Expectation Maximization algorithm. 

In \cite{Reed2014} the authors utilize what they call "perceptual consistency". They argue this is implicit in the network parameters and that it holds an internal representation of the world. Thus, it can serve as a basis for the network to "disagree" with the provided labels and relabel data during training. The network then "bootstraps" itself in this way, using what it learns from the relabeling as a basis to relabel more data, and so on.

These techniques are very robust to label noise in the image recognition but label errors in semantic segmentation present a fundamentally different problem, since label errors overwhelmingly occur at region boundaries, and no such concept exists for holistic image analysis. In addition, learning in semantic segmentation is done with fixed cohorts of pixels (images) within random batches. Therefore, a DL model may learn a general rule about feasible region size and discourage an otherwise positive prediction for a pixel in the absence of positive predictions for its neighbors.

\section{Methods}

\subsection{Perturbations}
\begin{figure}[H]
    \centering
    \includegraphics[width=12cm]{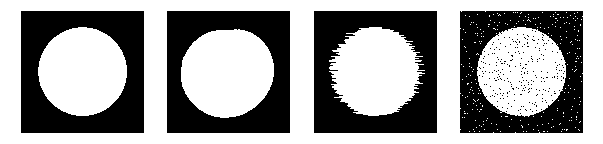}
    \caption{From left to right: unperturbed, natural perturbations, choppy perturbations, and random perturbations.}
    \label{fig:Perturbations}
\end{figure}

We attempted to perturb ground truth masks such that they closely mimicked the sorts of errors that human experts often make when drawing freehand contours. In order to achieve this, we first retrieved the contours from an existing binary mask using OpenCV's \verb+findContours()+ function. We then sampled points from this contour and moved them a random offset either towards or away from the contour's center. We used a simple fill to produce the perturbed annotation. The offsets were produced by a normal distribution with a given variance and zero mean. We call these offsets \textit{natural} perturbations. A natural perturbation applied to a circle can be seen in Fig. \ref{fig:Perturbations} (middle-left).

In addition, we wanted to mimic the sort of errors that occur when natural errors are made in a single plane of a volume and data is viewed from an orthogonal plane, as seen in Fig. \ref{fig:dif_plane_view}. For this, we iterated over every row in the masks, found each block of consecutive positive labels, and shifted the block's starting and end points by some amount that was once again sampled from a normal distribution with zero mean and provided variance. We call these \textit{choppy} perturbations. A choppy perturbation applied to a circle can be seen in Fig. \ref{fig:Perturbations} (middle-right).

Finally, in order to simulate random errors in a classification setting, we randomly chose an equal proportion of voxels from both the negative and positive classes and flipped their values. We call these \textit{random} perturbations (Fig. \ref{fig:Perturbations}, right).

Three parameter settings were chosen for each perturbation mode in order to produce perturbed ground truth with 0.95, 0.90, and 0.85 Dice-Sorensen agreement with the original ground truth, i.e. we chose 9 total parameter setting. Each was tuned by randomly choosing 1000 slices and using bisection with the terminal condition that the upper and lower bounds each producing Dice scores within 0.005 of the target.

\subsection{Training}
We ran the experiments using the Keras \cite{chollet2015keras} framework with a TensorFlow \cite{abadi2016tensorflow} back-end. We optimized our models using the Adam algorithm \cite{kingma2014adam} with the default parameter values. We addressed the imbalance of the problem by equally sampling from each class, and we used mini-batches of 20 slices, where each slice is a 512x512 array of Hounsfield Units from the axial plane. For each model, we started with 6 initial convolutional kernels and the number doubled with each down-sampling. Each model was trained for 100 epochs with 35 steps per epoch.

For each architecture and perturbation pair, we trained five times in order to improve statistical power, resulting in 150 total training sessions.

\section{Results}
Our results show that the performance of each model steadily declined with the extent of boundary-localized perturbations, but that model performance was very robust to random perturbations. This suggests that flawed ground truth labels, particularly in border regions, are hindering the performance of DL-based models for semantic segmentation. 

\begin{figure}[H]
    \centering
    \includegraphics[scale=0.52]{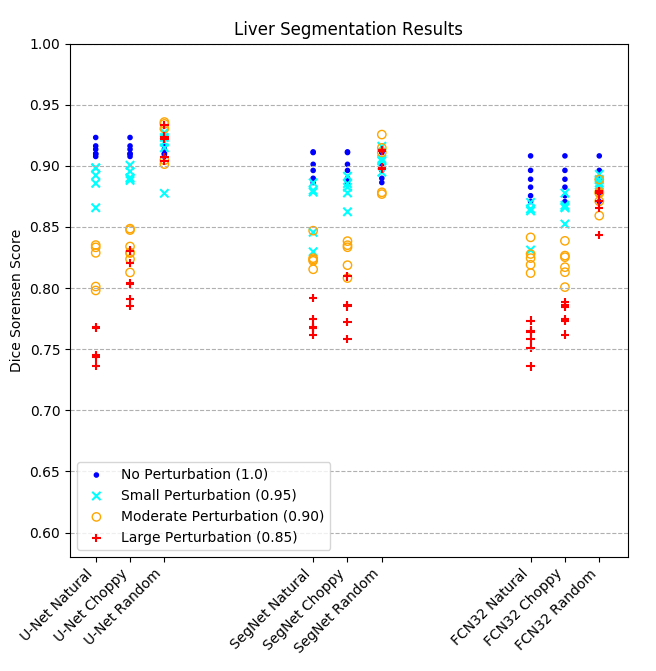}
    \caption{The results for liver segmentation for each model with each type of mode of ground-truth perturbation (best viewed in color).}
    \label{fig:lis_scatter}
\end{figure}

As can be seen in Fig. \ref{fig:lis_scatter}, other than the large choppy perturbations for U-Net, the responses of each architecture to the different degrees of boundary-localized perturbation were surprisingly uniform. This suggests that there may be a general predictive relationship between the incidence of ground-truth errors and the expected performance of these models. Additionally, it appears that each of the models are very resilient to random perturbations in ground truth, in some cases outperforming the Dice-Sorenson score of the training data itself by more than 5\%. 

\begin{table}
\begin{center}
\begin{tabu}{X[c]||X[c]|X[c]|X[c]}
     & U-Net & SegNet & FCN32\\
     \hline\hline
     Control & 0.9134 & 0.8993 & 0.8870\\
     \hline  
     Natural 0.95 & 0.8880 & 0.8640 & 0.8587\\
     \hline  
     Natural 0.90 & 0.8193 & 0.8265 & 0.8250\\
     \hline  
     Natural 0.85 & 0.7521 & 0.7717 & 0.7581\\
     \hline  
     Choppy 0.95 & 0.8928 & 0.8799 & 0.8660\\
     \hline  
     Choppy 0.90 & 0.8321 & 0.8268 & 0.8202\\
     \hline  
     Choppy 0.85 & 0.8058 & 0.7823 & 0.7782\\
     \hline  
     Random 0.95 & 0.9124 & 0.9050 & 0.8881\\
     \hline  
     Random 0.90 & 0.9213 & 0.9013 & 0.8751\\
     \hline  
     Random 0.85 & 0.9182 & 0.9068 & 0.8676\\
\end{tabu}
\vspace{0.5em}
\caption{Mean Dice-Sorensen score for each model-perturbation pair for Liver Segmentation}
\end{center}
\end{table}

U-Net's anomalously good performance in the presence of large choppy perturbations is interesting. We hypothesize that this is because U-Net's "skip connections" allow it to very effectively preserve border information from activation functions early on in the network. Thus, borders are likely still emphasized because even though the contour has become jagged, the region edges are centered on the true contour. This is not the case for the "natural" perturbations.

\section{Limitations}

Better performance has been reported for the liver segmentation problem \cite{le2016liver}, but that is due to the use of ensembles and hyperparameter tuning. It would not be feasible to engineer and train such techniques for each and every data point. It is possible (although we believe unlikely) that these findings do not translate to large ensemble settings, but this must be the subject for future work.

Additionally, these experiments were all run on a single dataset with binary labels. More work must be done to study whether these results generalize to different problems, and problems with many class labels. 

Finally, this study did not examine the effect of \em biased \em labels, which are also likely to exist in semantic segmentation datasets. Our intuition is that the models will tend to exhibit the same bias as the expert, but it's unclear what the effect on performance would be when there are multiple experts, each with different biases. This, too, must be the subject for future work.

\section{Conclusion}
In this work we tested how three widely-used deep learning based models responded to various modes of errors in ground-truth labels for semantic segmentation of the liver in abdominal CT scans. We found that in general, these models each experience relatively uniform performance degradation with increased incidence of label errors, but that U-Net was especially robust to large amounts of "choppy" noise on the liver regions.

There are many opportunities to continue this work. In particular, we would like to expand the scope of this study to look also at how the hyperparameters of the architectures and training procedures affect its sensitivity. We also believe it would be useful to explore the effect of dataset size on sensitivity, since it's possible that models will have a more difficult time coping with noisy data when they have less data to look at. Finally, we plan to study how deep-learning-based architectures for semantic segmentation can be modified in order to be more robust to ground truth errors at region boundaries.

The code for our experiments has been made available at\\ \url{https://github.com/neheller/labels18}.

\section*{Acknowledgements}
Research reported in this publication was supported by the National Cancer Institute of the National Institutes of Health under Award Number R01CA225435. The content is solely the responsibility of the authors and does not necessarily represent the official views of the National Institutes of Health.

\bibliographystyle{splncs04}
\bibliography{main}

\end{document}